\documentclass[runningheads]{llncs}
\usepackage{graphicx}
\usepackage{tikz}
\usepackage{comment}
\usepackage{amsmath,amssymb} 
\usepackage{color}
\usepackage[caption=false]{subfig}

\usepackage[accsupp]{axessibility}  

\usepackage[pagebackref,breaklinks,colorlinks]{hyperref}

\usepackage{booktabs}
\usepackage{multirow}
\usepackage{enumitem}
\usepackage{array}
\usepackage{makecell}

\definecolor{mydarkblue}{rgb}{0,0.08,1}
\definecolor{mydarkgreen}{rgb}{0.02,0.6,0.02}
\definecolor{myred}{rgb}{1.0,0.0,0.0}

\begin{document}
\def\mA{\mathcal{A}}
\def\mB{\mathcal{B}}
\def\mC{\mathcal{C}}
\def\mD{\mathcal{D}}
\def\mE{\mathcal{E}}
\def\mF{\mathcal{F}}
\def\mG{\mathcal{G}}
\def\mH{\mathcal{H}}
\def\mI{\mathcal{I}}
\def\mJ{\mathcal{J}}
\def\mK{\mathcal{K}}
\def\mL{\mathcal{L}}
\def\mM{\mathcal{M}}
\def\mN{\mathcal{N}}
\def\mO{\mathcal{O}}
\def\mP{\mathcal{P}}
\def\mQ{\mathcal{Q}}
\def\mR{\mathcal{R}}
\def\mS{\mathcal{S}}
\def\mT{\mathcal{T}}
\def\mU{\mathcal{U}}
\def\mV{\mathcal{V}}
\def\mW{\mathcal{W}}
\def\mX{\mathcal{X}}
\def\mY{\mathcal{Y}}
\def\mZ{\mathcal{Z}} 

\def\bbN{\mathbb{N}} 
\def\bbR{\mathbb{R}} 
\def\bbP{\mathbb{P}} 
\def\bbQ{\mathbb{Q}} 
\def\bbE{\mathbb{E}}

\def\1n{\mathbf{1}_n}
\def\0{\mathbf{0}}
\def\1{\mathbf{1}}

\def\A{{\bf A}}
\def\B{{\bf B}}
\def\C{{\bf C}}
\def\D{{\bf D}}
\def\E{{\bf E}}
\def\F{{\bf F}}
\def\G{{\bf G}}
\def\H{{\bf H}}
\def\I{{\bf I}}
\def\J{{\bf J}}
\def\K{{\bf K}}
\def\L{{\bf L}}
\def\M{{\bf M}}
\def\N{{\bf N}}
\def\O{{\bf O}}
\def\P{{\bf P}}
\def\Q{{\bf Q}}
\def\R{{\bf R}}
\def\S{{\bf S}}
\def\T{{\bf T}}
\def\U{{\bf U}}
\def\V{{\bf V}}
\def\W{{\bf W}}
\def\X{{\bf X}}
\def\Y{{\bf Y}}
\def\Z{{\bf Z}}

\def\a{{\bf a}}
\def\b{{\bf b}}
\def\c{{\bf c}}
\def\d{{\bf d}}
\def\e{{\bf e}}
\def\f{{\bf f}}
\def\g{{\bf g}}
\def\h{{\bf h}}
\def\i{{\bf i}}
\def\j{{\bf j}}
\def\k{{\bf k}}
\def\l{{\bf l}}
\def\m{{\bf m}}
\def\n{{\bf n}}
\def\o{{\bf o}}
\def\p{{\bf p}}
\def\q{{\bf q}}
\def\r{{\bf r}}
\def\s{{\bf s}}
\def\t{{\bf t}}
\def\u{{\bf u}}
\def\v{{\bf v}}
\def\w{{\bf w}}
\def\x{{\bf x}}
\def\y{{\bf y}}
\def\z{{\bf z}}

\def\balpha{\mbox{\boldmath{$\alpha$}}}
\def\bbeta{\mbox{\boldmath{$\beta$}}}
\def\bdelta{\mbox{\boldmath{$\delta$}}}
\def\bgamma{\mbox{\boldmath{$\gamma$}}}
\def\blambda{\mbox{\boldmath{$\lambda$}}}
\def\bsigma{\mbox{\boldmath{$\sigma$}}}
\def\btheta{\mbox{\boldmath{$\theta$}}}
\def\bomega{\mbox{\boldmath{$\omega$}}}
\def\bxi{\mbox{\boldmath{$\xi$}}}
\def\bnu{\mbox{\boldmath{$\nu$}}}                                  
\def\bphi{\mbox{\boldmath{$\phi$}}}
\def\bmu{\mbox{\boldmath{$\mu$}}}

\def\bDelta{\mbox{\boldmath{$\Delta$}}}
\def\bOmega{\mbox{\boldmath{$\Omega$}}}
\def\bPhi{\mbox{\boldmath{$\Phi$}}}
\def\bLambda{\mbox{\boldmath{$\Lambda$}}}
\def\bSigma{\mbox{\boldmath{$\Sigma$}}}
\def\bGamma{\mbox{\boldmath{$\Gamma$}}}
                                  
\newcommand{\myprob}[1]{\mathop{\mathbb{P}}_{#1}}

\newcommand{\myexp}[1]{\mathop{\mathbb{E}}_{#1}}

\newcommand{\mydelta}[1]{1_{#1}}

\newcommand{\myminimum}[1]{\mathop{\textrm{minimum}}_{#1}}
\newcommand{\mymaximum}[1]{\mathop{\textrm{maximum}}_{#1}}    
\newcommand{\mymin}[1]{\mathop{\textrm{minimize}}_{#1}}
\newcommand{\mymax}[1]{\mathop{\textrm{maximize}}_{#1}}
\newcommand{\mymins}[1]{\mathop{\textrm{min.}}_{#1}}
\newcommand{\mymaxs}[1]{\mathop{\textrm{max.}}_{#1}}  
\newcommand{\myargmin}[1]{\mathop{\textrm{argmin}}_{#1}} 
\newcommand{\myargmax}[1]{\mathop{\textrm{argmax}}_{#1}} 
\newcommand{\myst}{\textrm{s.t. }}

\newcommand{\denselist}{\itemsep -1pt}
\newcommand{\sparselist}{\itemsep 1pt}

\definecolor{pink}{rgb}{0.9,0.5,0.5}
\definecolor{purple}{rgb}{0.5, 0.4, 0.8}   
\definecolor{gray}{rgb}{0.3, 0.3, 0.3}
\definecolor{mygreen}{rgb}{0.2, 0.6, 0.2}

\newcommand{\cyan}[1]{\textcolor{cyan}{#1}}
\newcommand{\red}[1]{\textcolor{red}{#1}}  
\newcommand{\blue}[1]{\textcolor{blue}{#1}}
\newcommand{\magenta}[1]{\textcolor{magenta}{#1}}
\newcommand{\pink}[1]{\textcolor{pink}{#1}}
\newcommand{\green}[1]{\textcolor{green}{#1}} 
\newcommand{\gray}[1]{\textcolor{gray}{#1}}    
\newcommand{\mygreen}[1]{\textcolor{mygreen}{#1}}    
\newcommand{\purple}[1]{\textcolor{purple}{#1}}       

\definecolor{greena}{rgb}{0.4, 0.5, 0.1}
\newcommand{\greena}[1]{\textcolor{greena}{#1}}

\definecolor{bluea}{rgb}{0, 0.4, 0.6}
\newcommand{\bluea}[1]{\textcolor{bluea}{#1}}
\definecolor{reda}{rgb}{0.6, 0.2, 0.1}
\newcommand{\reda}[1]{\textcolor{reda}{#1}}

\def\changemargin#1#2{\list{}{\rightmargin#2\leftmargin#1}\item[]}
\let\endchangemargin=\endlist
                                               
\newcommand{\cm}[1]{}

\newcommand{\mhoai}[1]{{\color{magenta}\textbf{[MH: #1]}}}

\newcommand{\mtodo}[1]{{\color{red}$\blacksquare$\textbf{[TODO: #1]}}}
\newcommand{\myheading}[1]{\vspace{1ex}\noindent \textbf{#1}}
\newcommand{\htimesw}[2]{\mbox{$#1$$\times$$#2$}}


\newif\ifshowsolution
\showsolutiontrue

\ifshowsolution  
\newcommand{\Comment}[1]{\paragraph{\bf $\bigstar $ COMMENT:} {\sf #1} \bigskip}
\newcommand{\Solution}[2]{\paragraph{\bf $\bigstar $ SOLUTION:} {\sf #2} }
\newcommand{\Mistake}[2]{\paragraph{\bf $\blacksquare$ COMMON MISTAKE #1:} {\sf #2} \bigskip}
\else
\newcommand{\Solution}[2]{\vspace{#1}}
\fi

\newcommand{\truefalse}{
\begin{enumerate}
	\item True
	\item False
\end{enumerate}
}

\newcommand{\yesno}{
\begin{enumerate}
	\item Yes
	\item No
\end{enumerate}
}

\newcommand{\Sref}[1]{Sec.~\ref{#1}}
\newcommand{\Eref}[1]{Eq.~(\ref{#1})}
\newcommand{\Fref}[1]{Fig.~\ref{#1}}
\newcommand{\Tref}[1]{Table~\ref{#1}}

\pagestyle{headings}
\mainmatter
\def\ECCVSubNumber{5123}  
\def\Approach{Geodesic-Former}

\title{\Approach: a Geodesic-Guided Few-shot 3D Point Cloud Instance Segmenter}

\titlerunning{\Approach} 
\authorrunning{T. Ngo and K. Nguyen} 
\author{Tuan Ngo and Khoi Nguyen}
\institute{VinAI Research}

\maketitle

\begin{abstract}
This paper introduces a new problem in 3D point cloud: few-shot instance segmentation. Given a few annotated point clouds exemplified a target class, our goal is to segment all instances of this target class in a query point cloud. This problem has a wide range of practical applications where point-wise instance segmentation annotation is prohibitively expensive to collect. To address this problem, we present \Approach~-- the first geodesic-guided transformer for 3D point cloud instance segmentation. The key idea is to leverage the geodesic distance to tackle the density imbalance of LiDAR 3D point clouds. The LiDAR 3D point clouds are dense near the object surface and sparse or empty elsewhere making the Euclidean distance less effective to distinguish different objects. The geodesic distance, on the other hand, is more suitable since it encodes the scene's geometry which can be used as a guiding signal for the attention mechanism in a transformer decoder to generate kernels representing distinct features of instances. These kernels are then used in a dynamic convolution to obtain the final instance masks. To evaluate \Approach~on the new task, we propose new splits of the two common 3D point cloud instance segmentation datasets: ScannetV2 and S3DIS. \Approach~consistently outperforms strong baselines adapted from state-of-the-art 3D point cloud instance segmentation approaches with a significant margin. The code is available at \url{https://github.com/VinAIResearch/GeoFormer}.

\end{abstract}

\section{Introduction}
\label{sec:intro}
\begin{figure}[t]
  \centering
  \includegraphics[width=0.9\linewidth]{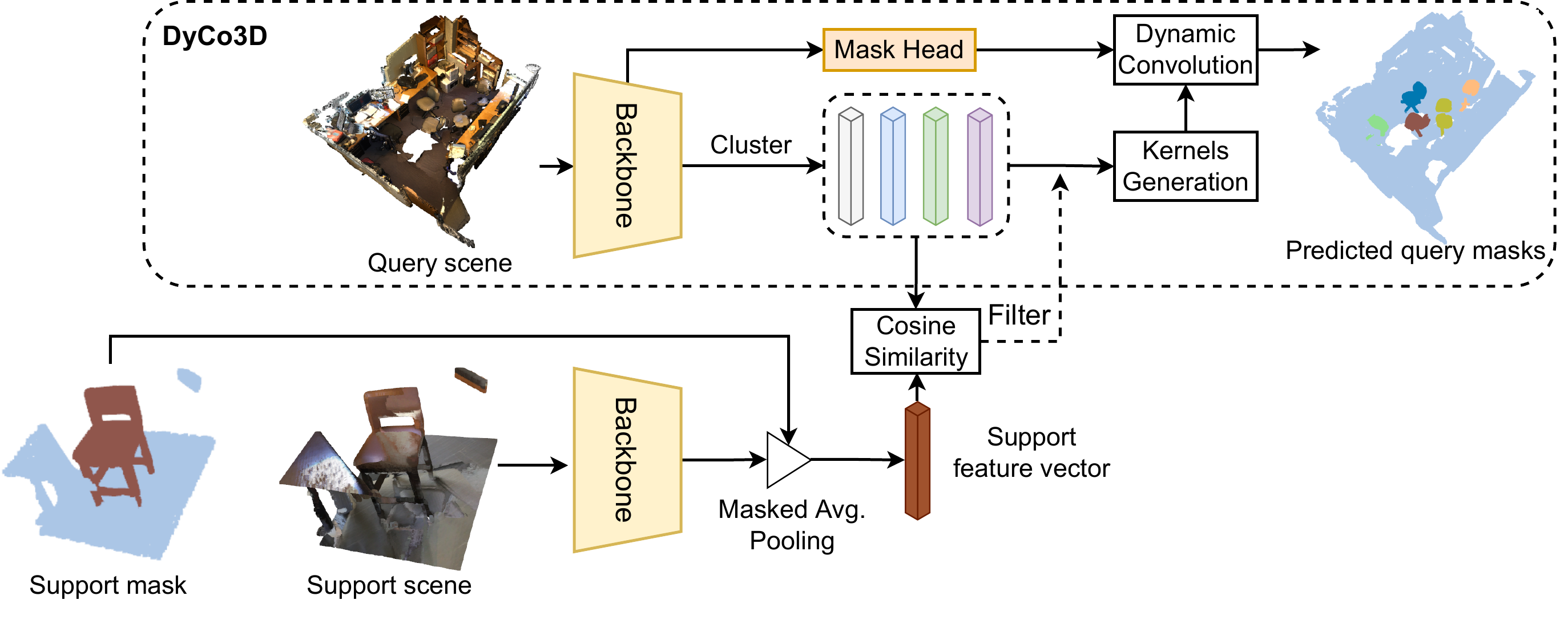}
  \vspace{-10pt}
   \caption{Our baseline adapted on DyCo3D \cite{he2021dyco3d} for 3DFSIS. The query and support point clouds are first input to a shared backbone to extract their features. Then the query points are grouped into candidates based on their semantic and predicted object centroids while the support points are masked-average-pooled to obtain a support feature vector.
   The cosine similarity between the support feature vector and every candidate's average feature is used to filter out irrelevant candidates. The final candidates are used to generate kernels for dynamic convolution with the feature produced by the mask head in order to obtain the final instance masks of the query scene.}
   \label{fig:baseline}
  \vspace{-10pt}
\end{figure}

This paper introduces a new problem of few-shot 3D point cloud instance segmentation (3DFSIS). As Fig.~\ref{fig:baseline} shows,
given a few support point clouds (a.k.a. scenes) with their ground-truth masks to define a target class, we aim to segment all instances of the target class in a query scene. 
%
Compared to related vision tasks such as 3D point cloud instance segmentation (3DIS) and 3D point cloud few-shot semantic segmentation (3DF3S), 3DFSIS is fundamentally different. 
For 3DIS, the training and test classes are the same. One could reliably learn an instance segmenter with abundant annotated examples in training, then apply that segmenter to the test scenes. That is not the case in 3DFSIS where training and test classes are disjoint.
For 3DF3S, we need to predict each point with a semantic label instead of an instance label as in 3DFSIS. That is, we do not need to distinguish different instances of the same class as in 3DF3S. 
Furthermore, unlike weakly/semi-supervised learning in 3DIS, where all classes are known in training, in the training of 3DFSIS, the new classes are not known in advance. Thus, the model needs to quickly learn from a few examples of new classes whenever they arrive.

3DFSIS is an important vision task and has a wide range of applications including autonomous driving, and augmented reality, especially in applications where training a reliable 3D instance segmenter is prohibitively impossible due to the expensive costs of collecting a sufficient amount of annotated point clouds. 
However, learning in 3D point clouds is very challenging due to: (1) 3D point clouds are unordered, imbalanced in density (dense near object surface but sparse elsewhere); and (2) the variance in appearance, size, and shape between the support and query scenes is significantly higher than that of 2D images. These two challenges are amplified in the few-shot setting where a very limited number of labeled examples of new classes are provided, e.g. 1 to 5 shots at most compared to 30 to 50 shots with ease in a 2D image. This is due to the reason that one has to label point-by-point in a 3D point cloud rather than labeling approximate polygons for instance masks as in a 2D image. Therefore, it is not trivial to adapt any 2DFSIS to 3DFSIS.

\begin{figure*}[t]
  \centering
  \includegraphics[width=1\linewidth]{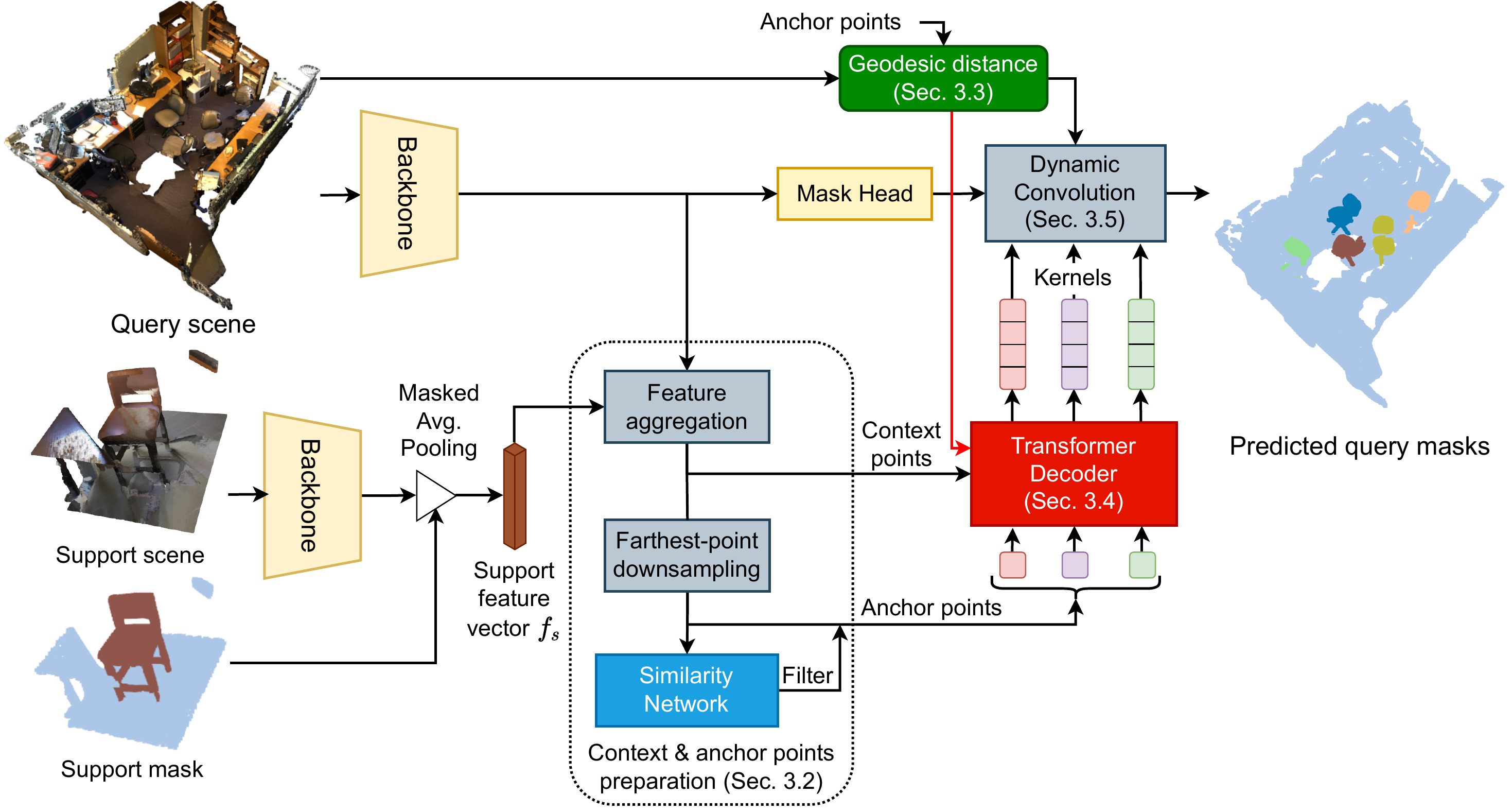}
  \vspace{-10pt}
   \caption{\textbf{Our proposed approach, \Approach, for 3DFSIS}. Given support and query scenes, a shared backbone is used to extract their features. Support features are further masked average pooled with the support mask to obtain a support feature vector representing the target class.
   Then the query features and support feature vector are aggregated to obtain \textit{context points} which are further sub-sampled in farthest-point-downsampling operation to obtain \textit{anchor points} representing the initial prediction of the object location.
    Next, the geodesic distances between every anchor point to all the context points are computed taking into account the imbalance in point cloud density (distributed near object surface only). This geodesic distance is used as the positional encoding to guide the transformer decoder whose key/value and query are context and anchor points, respectively, so as to produce a kernel for each anchor point. 
  Finally, each kernel dynamically convolves with the features produced by the mask head along with the geodesic distance embedding to obtain the final object instance mask.
   }
   \label{fig:architecture}
  \vspace{-10pt}
\end{figure*}

A simple but strong baseline for 3DFSIS can be adapted from a 3D point cloud instance segmenter, e.g. DyCo3D \cite{he2021dyco3d}, to the few-shot setting. The baseline is depicted in Fig.~\ref{fig:baseline}. 
First, similar points are grouped into candidates based on their Euclidean centroids and semantic predictions. Then each candidate is passed to a subnetwork to generate a kernel for dynamic convolution \cite{tian2020conditional} so as to obtain the final instance mask. 
To filter out the irrelevant candidates which do not belong to the target class, one can use cosine similarity between the support feature vector and the average feature vector of all points of each candidate.  
This framework has several limitations. First, as mentioned above, 3D point clouds are imbalanced in density and mostly distributed near the object surface so that the Euclidean distance for clustering is unreliable, i.e. points that are close together might not necessarily belong to the same object and vice versa. Second, the clustering in DyCo3D relies heavily on the performance of the offset centroid predictions, hence, it might be overfitting to some 3D shapes and sizes of the training object classes, resulting in poor generalization to the test classes. 



%

To address these limitations, we propose a new geodesic-guided transformer decoder to generate the kernel for the dynamic convolution from a set of initial anchor points, giving the name of our approach, \Approach. The overview of \Approach~is depicted in Fig.~\ref{fig:architecture}. 
First, geodesic distance embedding based on the geodesic distances between each of the anchor points to all context points is computed. In this way, the geodesic distance between two points belonging to different objects is very large, helping differentiate different objects.
Then this embedding is used as positional encoding to guide the later transformer decoder and dynamic convolution.
Second, to avoid overfitting to the shape and size of training classes, we use a combination of the Farthest Point Sampling \cite{qi2018pointnet}, a similarity network, and a transformer decoder. The first samples initial seeds from the query point cloud representing the initial locations of the object candidates, the second filters out irrelevant candidates, and the third contextualizes the foreground (FG) candidates to precisely represent objects with the information of the context points in order to generate the convolution kernels. In this way, as long as an initial seed belongs to an object, it can represent that object. In contrast, for each point, DyCo3D has to predict exactly the center point of the object it belongs to in order for the clustering to work well. This is even harder when transferring to the new object classes in testing. Also, to the best of our knowledge, we are the first to adopt the transformer decoder architecture to the 3DIS and 3DFSIS.


In sum, our technical contributions are summarized as follows:
\vspace{-5pt}
\begin{itemize}
    \item We introduce a new 3D point cloud few-shot instance segmentation task.
    \item To evaluate the new task, we introduce new splits adapted from the ScannetV2 and S3DIS datasets.
    \item To address the new task, we propose a strong baseline (adapted from SOTA 3DIS methods) and our novel proposed approach, \Approach, combining the transformer decoder with dynamic convolution in respect of the geodesic distance encoding scene's geometry.
\end{itemize}
\vspace{-5pt}
In the following, Sec.~\ref{sec:related_work} reviews prior work; Sec.~\ref{sec:approach} specifies \Approach; and Sec.~\ref{sec:experiments} presents our implementation details and experimental results. Sec.~\ref{sec:conclusion} concludes with some remarks and discussions.

\section{Related Work}
\label{sec:related_work}
This section reviews closely related work in 2D and 3D instance segmentation.

\myheading{3D point cloud instance segmentation (3DIS)} approaches can be divided into two groups: proposal-based and proposal-free. 
The \textit{proposal-based} approaches \cite{hou20193d,yang2019learning,yi2019gspn} first detect 3D bounding boxes, then segment the foreground region inside them. 3D-SIS \cite{hou20193d} proposes a Mask R-CNN-based 3D instance segmentation architecture, jointly learns features from both RGB images and 3D point cloud. 
3D-BoNet \cite{yang2019learning} simplifies the detection network by directly predicting a fixed number of unoriented 3D bounding boxes from a global feature vector, then segmenting foreground points inside each box. GSPN \cite{yi2019gspn} generates proposals by reconstructing shapes from noisy observations and further refining these proposals with a Region-based PointNet \cite{qi2017pointnet}. 
On the other hand, the \textit{proposal-free} approaches \cite{wang2018sgpn,jiang2020pointgroup,chen2021hierarchical,he2021dyco3d,engelmann20203d} learn embedding features then group points to instances. SGPN \cite{wang2018sgpn} adopts the double-hinge loss to learn discriminative features in order to compute the similarity matrix of paired points for grouping points. PointGroup \cite{jiang2020pointgroup} predicts the 3D offset from each point to its instance's centroid and generates clusters from two sets: original points and shifted points. HAIS \cite{chen2021hierarchical} proposes a hierarchical clustering method where a small cluster can be either filtered out or absorbed by a larger cluster to become its part. DyCo3D \cite{he2021dyco3d} adopts the same clustering approach but leverages dynamic convolution \cite{wang2020solov2,tian2020conditional} to generate 3D instance masks. SSTNet \cite{liang2021instance} constructs a super point tree based on the point cloud's semantic features and uses tree traversal to split nodes into instances.
All the 3DIS approaches assume the training and test classes are the same, and there is a large number of annotated data for training. The setting of 3DFIS is fundamentally different: the training and test classes are disjoint and we only have a few annotated examples for each test class. 

\myheading{Few-shot 2D instance segmentation} approaches \cite{michaelis2018one,yan2019meta,fan2020fgn,nguyen2021fapis,nguyen2022ifs} extend the Mask R-CNN\cite{he2017mask} -- a common 2D image instance segmenter -- to the few-shot setting. The support features are extracted from a few labeled examples and incorporated into the query feature map to segment objects of the target class.
\cite{nguyen2021fapis} utilizes the anchor-free detector \cite{tian2019fcos} to alleviate the overfitting problem of the anchor boxes to the training classes and assembles the predicted object's latent parts into an object mask.
However, 2D images are structured, grid-based, and dense whereas 3D point clouds are unordered, irregular, and sparse. Therefore, these approaches cannot be applied directly to 3DFSIS.

\myheading{Few-shot 3D point cloud semantic segmentation (3DF3S)}. Recently, \cite{zhao2021few} introduced the problem of few-shot 3D point cloud semantic segmentation. From the support scene, multiple prototypes are extracted and propagated to the query points based on their affinity matrix. However, this approach does not distinguish different instances of the same object class. 3DFSIS is arguably harder than 3DFSSS since we need to classify all points into instance labels instead of semantic labels only. 

\myheading{Vision transformer} has been applied to 2D image classification \cite{dosovitskiy2020image,touvron2021training,yuan2021tokens}, object detection \cite{carion2020end,zhu2020deformable,wang2021pyramid,fang2021you,meng2021conditional,wang2021anchor}, semantic segmentation \cite{xie2021segformer,strudel2021segmenter}, and instance segmentation \cite{li2021panoptic,guo2021sotr,dong2021solq,hu2021istr}. 
Furthermore, the transformer architecture is naturally fit to process unordered 3D point clouds since its attention mechanism is permutation invariant. Some recent approaches \cite{liu2020tanet,zhao2021point,misra2021end} have shown the potential of transformers in some 3D tasks.
\cite{zhao2021point} designs a self-attention network to process 3D point clouds and achieve good results on 3D semantic segmentation, object part segmentation, and object classification. \cite{liu2020tanet,pan20213d,misra2021end} leverage transformer-based architecture for 3D object detection. 
We are the first to adopt the cross-attention transformer decoder with a special design for the 3DIS and 3DFSIS tasks.

\section{Our \Approach}
\label{sec:approach}

\subsection{Problem setting} 

In training, we are provided a sufficiently large training set of base classes $C_{train}$, i.e. $\{P^t, m^t\}_{t=1}^{T}$, where $P^t, m^t$ are the 3D point cloud of the scene $t$ and its ground-truth segmentation masks, respectively, and $T$ is the number of training samples.
In testing, given $K$ support 3D point cloud scenes $P_s$ and their ground-truth masks $m_s$ to define a new target class $c_{test}$, we seek to segment all instances $m_q$ of the target class in a query scene $P_q$. It is worth noting that the target class is different from the base classes, or $c_{test} \notin C_{train}$. 
In this paper, we explore two configurations: 1-shot and 5-shot instance segmentation.

To address this problem, we design our approach \Approach~inspired by DyCo3D \cite{he2021dyco3d}. 
The overview of \Approach~is illustrated in Fig.~\ref{fig:architecture}. 
To extract features $F_s, F_q$ from the support and query point clouds $P_s, P_q$, respectively, we employ a 3D U-Net with sparse convolution \cite{graham20183d} used in \cite{he2021dyco3d}.
In addition, a support feature vector $f_s$ is extracted from the support features $F_s$ via a masked-average-pooling operation representing the target class.

In the following, Sec.~\ref{sec:preparation} first describes how to prepare the context and anchor points for the transformer decoder. Sec.~\ref{sec:geodesic} specifies how to compute the geodesic distance between each anchor point to every context point to guide the transformer decoder and dynamic convolution.
Sec.~\ref{sec:decoder} discusses how the transformer decoder generates the convolution kernel for the dynamic convolution, which is presented in Sec.~\ref{sec:dyco}, in order to produce the final instance mask.
Finally, Sec.~\ref{sec:training_strategy} proposes the strategy to train our approach.


\subsection{Context and Anchor Points Preparation}
\label{sec:preparation}

First, we aggregate support feature $f_s\in \mathbb{R}^{1\times d}$ into the query features $F_q \in \mathbb{R}^{N_q \times d}$ inspired by \cite{xiao2020few}, resulting in integrated features of the \textit{context points} $F_{c} \in \mathbb{R}^{N_q \times d}$ as follows:
\begin{equation}
    F_c = W_{proj} * \left[F_{q} \odot f_{s}; F_{q} - f_{s}; F_{q}\right],
    \label{eq:aggregation}
\end{equation}
where $d$ is the number of feature channels, $N_q$ is the number of query points, $W_{proj} \in \mathbb{R}^{3d \times d}$ is the linear layer weight; $*, [\cdot;\cdot], \odot, -$ are the convolution, concatenation, channel-wise multiplication, and subtraction operations, respectively. In this way, we preserve the original query point features along with the newly rectified and subtracted features from the support.

Next, from the context points, a smaller set of points is sampled by a farthest-point-down sampling to represent distinct object candidates. In our work, we sample a large enough number of candidates so that they can cover all objects in all cases.
Then, a similarity network, which is a multi-layer perceptron (MLP), is used to filter relevant candidates as \textit{anchor points} $F_a \in \mathbb{R}^{N_a \times d}$, $N_a$ is the number of anchor points, having high appearance similarity with the support examples.
After this step, we take the context points $F_c$ and anchor points $F_a$ as input to the transformer decoder to generate the kernel of each anchor point. 


\subsection{Geodesic Distance Embedding Computation}
\label{sec:geodesic}
\begin{figure}[t]
  \subfloat[]{
	\begin{minipage}[l]{
	   0.4\textwidth}
	   \centering
	   \label{fig:euclid_geo}
	   \includegraphics[width=1\textwidth]{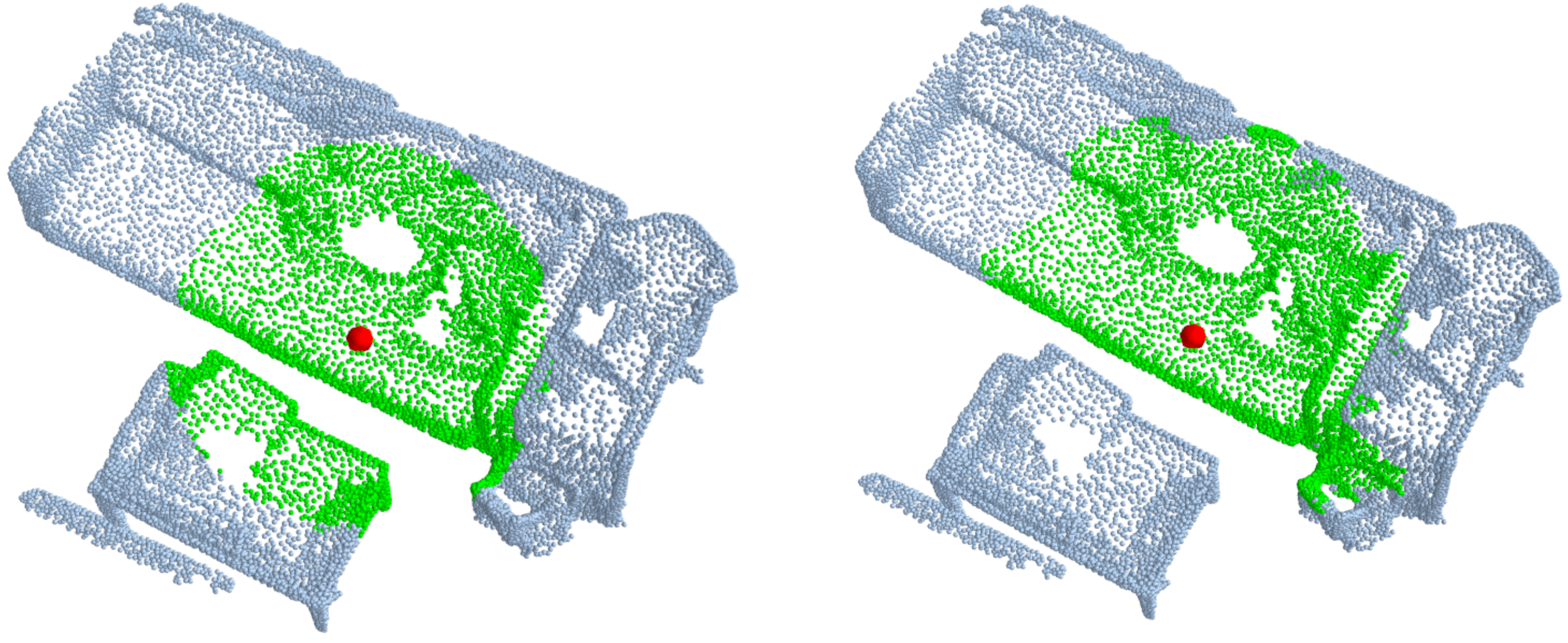}
	\end{minipage}}
  \hspace{25pt}
  \subfloat[]{
	\begin{minipage}[r]{
	   0.45\textwidth}
	   \centering
	   \label{fig:fg_bg}
	   \includegraphics[width=\textwidth]{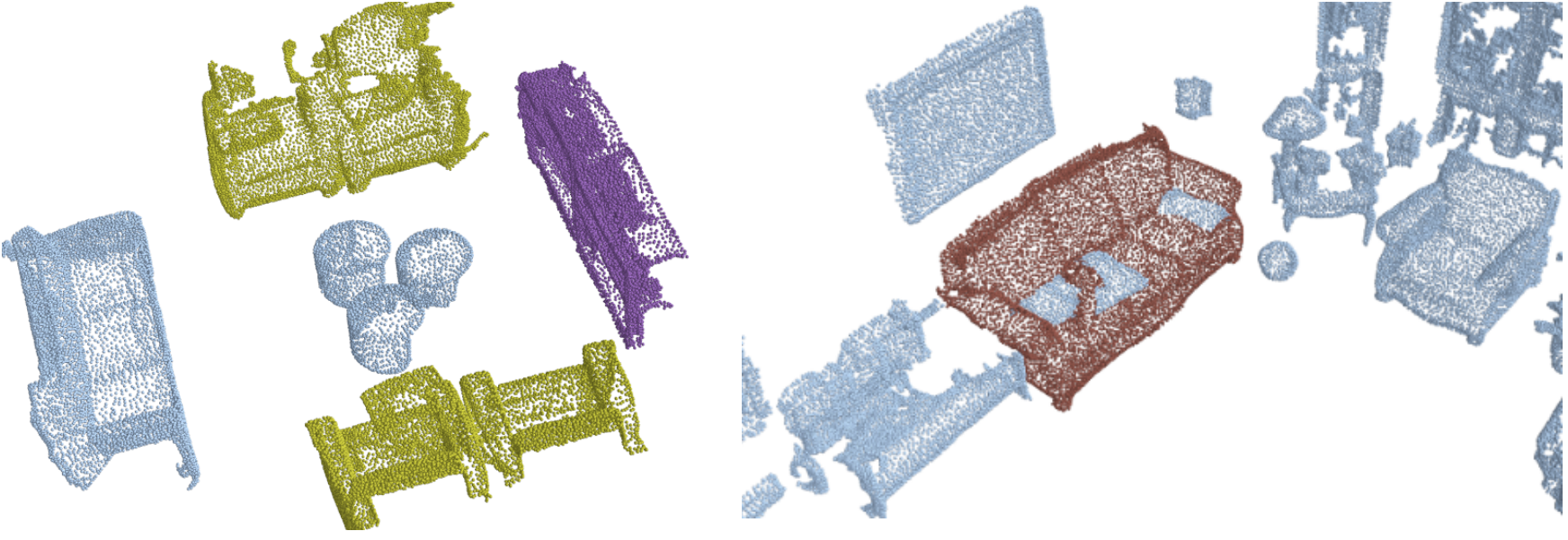}
	\end{minipage}}
\vspace{-10pt}
\caption{(a) Comparison between Euclidean distance and geodesic distance. For each image, the green points are the top-2000 nearest neighbors of the red point in Euclidean distance (left) and geodesic distance (right). (b) An example of FG/BG flipping in training and testing making transformer classifier confused, i.e. sofa is labeled as BG in training (left) but FG in testing (right).
}
\vspace{-10pt}
\end{figure}

The 3D point clouds captured by LiDAR sensors have an important property that it is distributed unequally in the 3D space (dense near the object surface and sparse elsewhere). As a result, two points are close in 3D Euclidean distance but they might belong to two different objects.
In this case, the geodesic distance \cite{kimmel1998computing} which encodes the scene's geometry would be a better choice  as visualized in Fig.~\ref{fig:euclid_geo}. In other words, if two points are close in Euclidean distance but there is no path or their geodesic distance is too high, they clearly belong to two separate objects. Therefore, we propose to use the geodesic distance between the anchor points and every context point as geometry guidance to distinguish objects in subsequent modules.  

To obtain the geodesic distance, we first employ the ball query algorithm \cite{qi2018pointnet} to get a directed sparse graph whose nodes are context points and each node only connects to at most $\kappa$ other nodes. There exists a directed edge from node 1 to node 2 if node 2 is among the $\kappa$ nearest neighbors of node 1 and within a radius $\tau$, and the weight of the edge is always positive and equal to the local Euclidean distance between the two nodes. After that, we use the shortest path algorithm, i.e. Djikstra \cite{dijkstra1959note}, to compute the length of the shortest path from each anchor point to every context point in the obtained sparse graph as its geodesic distance. Finally, the geodesic distance embedding $G^i \in \mathbb{R}^{N_q \times d}$ of an anchor point $i$ is obtained by encoding its geodesic distance using the sine/cosine function in \cite{vaswani2017attention}.

\subsection{Transformer Decoder}
\label{sec:decoder}


The transformer decoder takes as input the anchor points $F_a \in \mathbb{R}^{N_a \times d}$ and context points $F_c \in \mathbb{R}^{N_q \times d}$ to produce the kernel $W^i \in \mathbb{R}^{L}$ for each anchor point $i$,  where $L$ is the number of parameters in dynamic convolution.
The decoder follows the design of DETR \cite{carion2020end} consisting of a multi-block of transformer layers with two kinds of attention: self-attention between anchor points and cross-attention between anchor and context points. 
Hence, each anchor point knows each other and captures a complete object structure to generate a kernel for the dynamic convolution. 
Notably, the attention mechanism in a transformer is inherently fitted to the 3D point cloud since they are both unordered. 

Importantly, to address the 3DFSIS, we make substantial modifications to the positional encoding and output of the decoder. First, to guide the attention in the transformer with the geodesic geometry structure as discussed in Sec.~\ref{sec:geodesic}, the geodesic distance embedding $G$ is used as the positional encoding instead of the embedding of 3D point coordinates.
Second, we do not predict the object class, i.e., the foreground (FG)/background (BG) classification, due to the FG and BG confusion of few-shot settings during the training and testing phase. In particular, a lot of new classes presenting in training scenes but are labeled as BG causing the trained classification head to predict them as BG (false negative) in testing as depicted in Fig.~\ref{fig:fg_bg}. Instead, the similarity network filters the FG anchor points as described in Sec.~\ref{sec:preparation}. 

\subsection{Dynamic Convolution}
\label{sec:dyco}

To prepare features for dynamic convolution whose weights are predicted by the transformer decoder, a
\textbf{mask head} takes as input the query point features $F_q \in \mathbb{R}^{N_q \times d}$ to produce the mask features $F_{mask} \in \mathbb{R}^{N_q \times d}$ by applying two blocks of MLP with Batch Norm \cite{ioffe2015batch}, and ReLU \cite{nair2010rectified} in between. Also, the geodesic distance is critical geometric cue to distinguish instances, we directly append the geodesic distance embedding $G^i \in \mathbb{R}^{N_q \times d}$ of each anchor point $i$ to the mask features in order to obtain the final instance mask $\widehat{m}^{i} \in [0,1]^{N_q \times 1}$ in a \textbf{dynamic convolution} as follows :
\begin{equation}
    \widehat{m}^{i} = \text{Conv} \left( [F_{mask}; G^i]; W^i \right),
\end{equation}
where $[\cdot;\cdot]$ is the concatenation operation, and Conv is implemented with several convolutional layers as in DyCo3D \cite{he2021dyco3d}.





\subsection{Training Strategy}
\label{sec:training_strategy}

\textbf{Pretraining:} 
First, we pretrain the U-Net backbone, mask head, and the transformer decoder with the standard 3D point cloud instance segmentation task on the base classes. In this stage, the feature aggregation in Fig.~\ref{fig:architecture} is not used since we do not have support feature, instead, we copy the features of query points to the context points directly. Also, we add a classification head on top of the output of the transformer decoder to predict the semantic category $\widehat{\gamma}^i$ along with the kernel generation to predict the mask $\widehat{m}^i$ for each anchor point $i$. The number of classes is $\Gamma+1$ where $\Gamma$ is the total number of base classes of $C_{train}$ and one additional background class. The matching cost $C_{\text{match}}^{\text{pretrain}} \in \mathbb{R}_+^{N_a \times N_{gt}}$ between the prediction $(\widehat{\gamma}^i, \widehat{m}^i)$ and the ground truth $(\gamma^j, m^j)$ is computed as:
\begin{equation}
\label{eqn:matching_cost}
    C_{\text{match}}^{\text{pretrain}}(i, j) = L_{seg}(\widehat{m}^i,m^j) + L_{cls}(\widehat{\gamma}^i,\gamma^j), 
\end{equation}
where $L_{seg}$ is the dice loss \cite{sudre2017generalised}, and $L_{cls}$ is the sigmoid focal loss \cite{lin2017focal}. Based on the matching cost $C_{\text{match}}^{\text{pretrain}}$, the Hungarian algorithm \cite{kuhn1955hungarian} is leveraged to find the optimal 1-to-1 matching $\pi^*$, then the following loss is used for training:
\begin{equation}
\label{eqn:hungarian}
    L_{\text{Hungarian}}^{\text{pretrain}} = \sum_{i=1}^{N_{GT}}L_{seg}(\widehat{m}^{i}, m^{\pi^*(i)}) + \sum_{i=1}^{N_a} L_{cls}(\widehat{\gamma}^i,\gamma^{\pi^*(i)}).
\end{equation}
If a class prediction $\widehat{\gamma}^i$ has no ground truths matched, it will be matched with the background class.

\myheading{Episodic training:} We leverage the episodic training strategy -- a common approach for few-shot image classification -- to mimic the test scenario in training. That is, for each episode, we randomly sample a pair of support and query point clouds $P_{s}, P_{q}$ and their masks $m_s, m_q$ from training examples of the base classes.
In this stage, the classification head is removed and we add feature aggregation and similarity network to train with the transformer decoder while freezing the backbone and mask head. This is the final architecture of \Approach~as depicted in Fig.~\ref{fig:architecture}. The following matching cost and loss are used to train \ and Approach~in this stage:
\begin{align}
    C_{\text{match}}^{\text{episodic}}(i, j) = L_{seg}(\widehat{m}^i,m^j), \quad
    L_{\text{Hungarian}}^{\text{episodic}} = \sum_{j=1}^{N_{GT}}L_{seg}(\widehat{m}^{i}, m^{\pi^*(i)}).
\end{align}
\textbf{For $K>1$ shots}, we additionally apply the episodic training on a set of balanced support-query pairs of the base and new classes to further fine-tune the \Approach. In testing, the final support feature vector $f_s$ is the average vector of all feature vectors $f_s^k$ of $K$ support scenes.

\section{Experiments}
\label{sec:experiments}

\begin{table}[t]
\small
\setlength{\tabcolsep}{10pt}
\centering
\caption{Class splits of the ScannetV2 and S3DIS datasets. Fold 0 is used for training while fold 1 is used for testing.}
\vspace{-10pt}
\begin{tabular}{cc cc}
\toprule
\multicolumn{2}{c}{ScannetV2} & \multicolumn{2}{c}{S3DIS} \\
\cmidrule(lr){1-2} \cmidrule(lr){3-4}
Fold 0 & Fold 1 & Fold 0 & Fold 1 \\ \midrule
cabinet & sofa & beam & door \\
bed     & table & board & floor\\ 
chair   & window & bookcase & sofa\\
door    & picture & ceiling & table\\
bookshelf  & shower curtain & chair & wall \\
counter & refrigerator & column & window\\
desk    & toilet & & \\
curtain & sink & & \\
bathtub & other furniture & &  \\
\bottomrule
\end{tabular}
\label{tab:split_dataset}
\vspace{-10pt}
\end{table}

\textbf{Datasets:}
To evaluate \Approach~on the new 3DFSIS task, we introduce two new datasets derived from ScannetV2 \cite{dai2017scannet} and S3DIS \cite{armeni20163d} used for 3D point cloud instance segmentation.
ScannetV2 consists of 1613 point clouds of scans from 707 unique indoor scenes with 20 semantic classes in total and 18 classes for instance segmentation. We follow the common split of 1201, 312, and 100 for training, evaluating, and testing, respectively \cite{he2021dyco3d}. 
Inspired by \cite{zhao2021few} for 3D few-shot semantic segmentation, we split the 18 foreground classes into two non-overlapping folds based on the alphabetical order with nine classes each, one for training classes (fold 0) and the other for test classes (fold 1). 
S3DIS is another benchmark for 3D indoor scenes which contains 272 point clouds collected from 6 large-scale areas with 13 semantic categories. We only keep 12 main categories and remove the ``clutter" class. We also split it into two folds with six classes each. Area 5 containing 68 point clouds is used for testing while the rest is used for training. Tab.~\ref{tab:split_dataset} summarizes the class splits of ScannetV2 and S3DIS.

We report the results for the test classes in the following procedure:
(1) we randomly sample $K=\{1, 5\}$ support examples, with their binary masks for every class in the training set (with the purpose of saving the whole test set for the query scenes only) and apply them to the whole test set, a.k.a the fixed support set; (2) for each query scene in the test set, if a test class does not present in the scene, we skip the evaluation of that class for that scene. To improve the reliability of the measured metrics, we sample and evaluate all the approaches on ten disjoint fixed support sets, and report the average with standard deviation. In this setting, we consider the unlabeled points of new classes in the training set as unseen points commonly used in 2DFSIS.

\myheading{Evaluation metrics:} For ScannetV2, we adopt the mean average
precision (mAP) and AP50 used in the instance segmentation task. For S3DIS, we apply the metrics that are used in \cite{jiang2020pointgroup,he2021dyco3d,he2020learning,he2020instance} to test classes: mCov, mPrec, and mRec. They are the mean instance-wise IoU, mean precision, and mean recall.

\myheading{Implementation details:}
We adopt the sparse convolution \cite{graham20183d} to implement the backbone network. The voxel size is set to 0.02 m for ScannetV2 and 0.05 m for S3DIS, and the output channel of the backbone network is set to 16. 
To calculate the geodesic distance, we employ the FAISS\footnote{\url{https://github.com/facebookresearch/faiss}} library for ball-query search, then we re-implement by vectorizing the shortest path algorithm, i.e., Dijkstra's algorithm, in Pytorch to speed up the processing time. 
The transformer decoder is the same as \cite{misra2021end} consisting of four layers, each uses multi-head attention with four heads, and the output dimension and the hidden dimension are set to 64.
We train our model using the Adam optimizer \cite{kingma2014adam} with a cosine learning rate scheduler \cite{loshchilov2016sgdr}. During the pretraining phase, the initial learning rate is set to $10^{-2}$, and the number of training epochs is 500. After that, we train for another 200 epochs in episodic training with the learning rate of $5 \times 10^{-3}$. Our data augmentation is the same as \cite{jiang2020pointgroup}'s.

\begin{table}[t]
\small
\caption{Ablation study on each component's contribution to the final results. ``SN", ``TD", and ``GDE'' denote similarity network, transformer decoder, and geodesic distance embedding, respectively. (*) denotes the baseline of per-point classification.}
\vspace{-10pt}
\setlength{\tabcolsep}{10pt}
\centering
\begin{tabular}{lccccc}
\toprule
  & \multicolumn{3}{c}{Combination} & \multicolumn{2}{c}{Metric} \\
\cmidrule(lr){2-4} \cmidrule(lr){5-6} 
  & SN  &  TD  & GDE  & mAP  & AP50 \\ 
 \midrule
Baseline (DyCo3D \cite{he2021dyco3d})     &     &       &           & 6.2 & 11.7    \\ 
Baseline (*)     &     &       &           & 4.9 & 9.7    \\ \midrule
  &  \checkmark &  &      & 6.6 & 12.5    \\
  &       &  \checkmark     &      & 6.7 & 13.1    \\
  & & & \checkmark & 7.8 & 14.2 \\
  &  \checkmark & \checkmark  &       & 7.6 & 14.3   \\ 
  &  \checkmark &  & \checkmark     & 8.7 & 14.9     \\ 
  &  & \checkmark & \checkmark     & 9.4 & 17.1      \\ 
 \midrule
\Approach &  \checkmark & \checkmark  &  \checkmark    & \textbf{10.6} & \textbf{19.8} \\ 
\Approach~w/ cls.   & &  \checkmark  & \checkmark     & 4.5 & 10.2 \\ 
\bottomrule
\end{tabular}
\label{tab:accumimpact}
\vspace{-10pt}
\end{table}

\subsection{Ablation Study}

\begin{figure}[t]
    \centering
    \includegraphics[width=.5\linewidth]{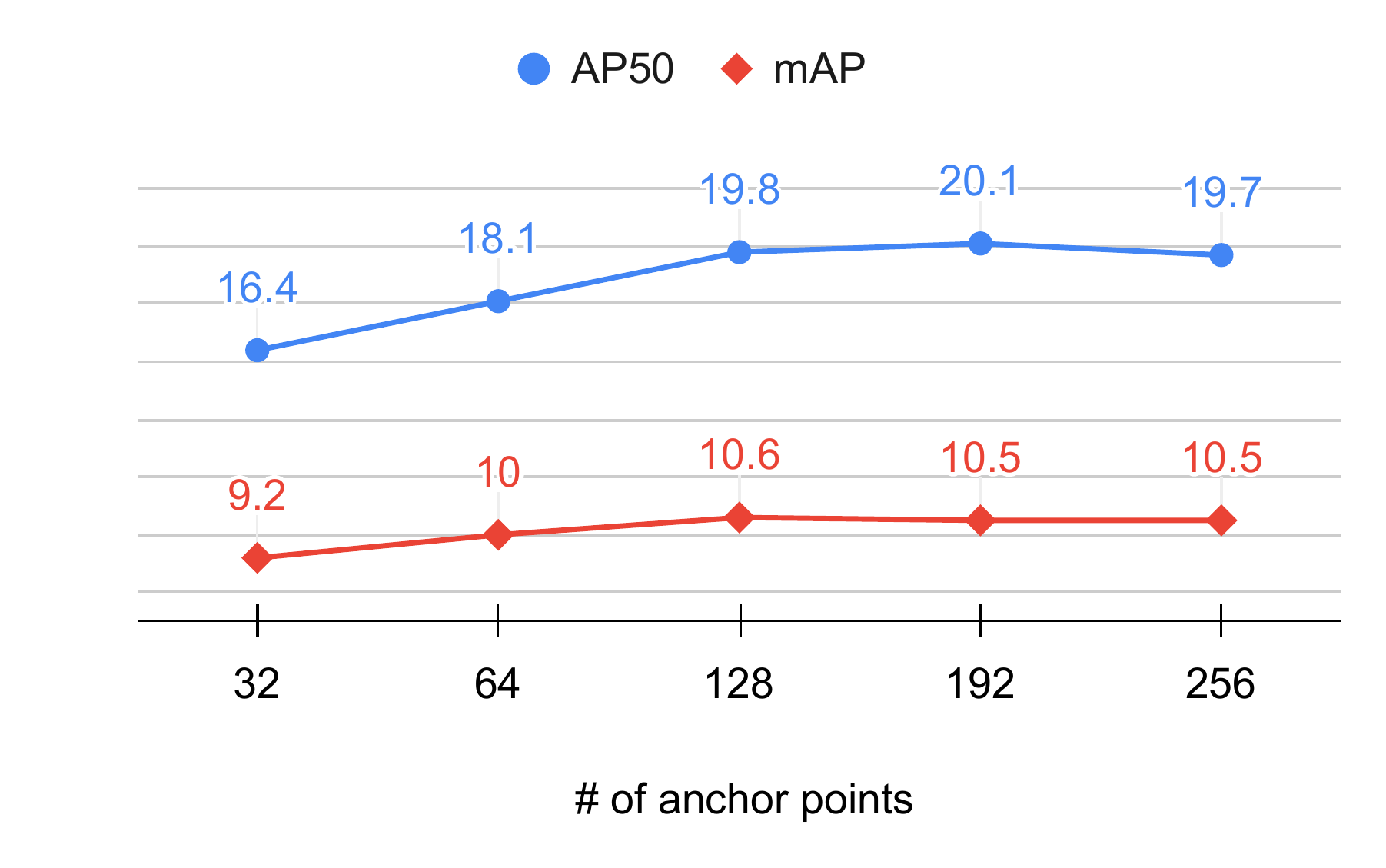}
    \vspace{-10pt}
    \caption{Study on the number of anchor points $N_a$.}
    \label{fig:num_queries_chart}
    \vspace{-10pt}
\end{figure}

\begin{table}[t] 
\caption{Study on the number of dynamic convolution layers.}
\vspace{-10pt}
\setlength{\tabcolsep}{8pt}
\centering
\begin{tabular}{lcccc}
\toprule
\# of layers & 1  & 2 & 3 & 4\\ \midrule
mAP on training set& 22.6 $\pm$ 1.4 & 28.1 $\pm$ 1.7 & 28.0 $\pm$ 1.3 & 28.3 $\pm$ 1.5\\
mAP on testing set& 3.4 $\pm$ 0.2 & 10.6 $\pm$ 0.6 & 9.3 $\pm$ 1.3  & 6.4 $\pm$ 1.9\\
\bottomrule
\end{tabular}
\label{tab:dyconv_overfit}
\end{table}

\begin{table}[t] 
\centering
\setlength{\tabcolsep}{8pt}
\caption{Study on ball query settings in Sec.~\ref{sec:geodesic} to form sparse directed graph.}
\vspace{-10pt}
\begin{tabular}{lccccc}
\toprule
mAP         & $\kappa=16$ & $\kappa=32$ & $\kappa=64$ & $\kappa=128$\\ \midrule
$\tau=0.03$ m  & 9.0 $\pm$ 0.9    & 9.3 $\pm$ 0.8  & 9.9 $\pm$ 0.7  & 10.1 $\pm$ 0.8\\
$\tau=0.05$ m  & 9.2 $\pm$ 0.6   & 9.7 $\pm$ 0.8   & \textbf{10.6 $\pm$ 0.6}  & 10.6 $\pm$ 0.7\\
$\tau=0.1$ m & 8.9 $\pm$ 1.2    & 9.3 $\pm$ 0.7    & 10.5 $\pm$ 1.0 & 10.3 $\pm$ 0.9\\ 
\bottomrule
\end{tabular}
\label{tab:ball_query_setting}
\vspace{-10pt}
\end{table}

We conduct several experiments on the validation set of ScannetV2 to study the contribution of various components of our method with one shot, $K=1$.

\myheading{Similarity network, transformer decoder, geodesic distance.} In Tab.~\ref{tab:accumimpact}, the first and second rows show the performance of our baseline in Fig.~\ref{fig:baseline}, and a per-point classification variant where we use cosine similarity to filter out irrelevant points before clustering by predicted objects' centers. This variant performs poorly as each point is classified independently without geometric cues of objects, and the classified points are so cluttered to form a complete shape. When replacing the cosine similarity in the baseline with a similarity network, the performance slightly increases, +0.4 in row 3. When the clustering algorithm in the baseline is replaced by the transformer decoder, the performance also slightly improves, +0.5 in row 4. Especially, when adding the geodesic distance embedding to the dynamic convolution of the baseline, the performance is significantly boosted, +1.6 in row 5. This justifies the importance of geodesic distance to the segmentation. 
When combining each pair of the three components, the performance improves substantially over each component alone. Finally, our full approach, \Approach~achieves the best performance, 10.6 in mAP and 19.8 in AP50. These results show that when combining these components together, the performance gain is much larger than using them separately.
We also have an ablation when turning off the similarity network and using the classification head in the pretraining phase, the performance drops significantly, -6.1 in row 9. This justifies our claims that using the classification head in our 3DFSIS is sub-optimal due to the FG/BG confusion as described in Sec.~\ref{sec:decoder}.

\myheading{Number of anchor points.} The results are summarized in Fig.~\ref{fig:num_queries_chart}. 
Using the number of anchor points of 128 gives the best results. This is because using too few anchor points cannot capture the diversity of objects in the scene, whereas using too many does not boost the performance significantly. 


\myheading{Number of layers in the dynamic convolution.} As can be seen in Tab.~\ref{tab:dyconv_overfit}, using only a single layer of dynamic convolution leads to a significant drop in performance (-7.2 in mAP). 
On the other hand, using too many layers may be prone to overfitting the training data and harder to optimize due to a large number of generated parameters. Using two layers gives the best results.

\myheading{Ball query configuration.}
Tab.~\ref{tab:ball_query_setting} reports the results with different nearest neighbors $\kappa$ and radii $\tau$ to form the directed sparse graph (as described in Sec.~\ref{sec:geodesic}) in order to compute the geodesic distance. From this table, $\kappa=64$ and $\tau=0.05$ m give the best results.

\begin{table}[t]
\small
\setlength{\tabcolsep}{8pt}
\caption{Comparison of \Approach~and the strong baselines on ScannetV2.}
\vspace{-10pt}
\centering
\begin{tabular}{lcccc}
\toprule
    &   \multicolumn{2}{c}{$K=1$}    & \multicolumn{2}{c}{$K=5$} \\ \cmidrule(lr){2-3}\cmidrule(lr){4-5}
            & mAP     & AP50 & mAP     & AP50 \\ 
\midrule
DyCo3D \cite{he2021dyco3d}            & 6.2 $\pm$ 2.0  & 11.7 $\pm$ 3.1 & 6.4 $\pm$ 1.2  & 11.9 $\pm$ 2.2     \\
PointGroup \cite{jiang2020pointgroup} & 5.3 $\pm$ 1.2  & 10.3 $\pm$ 2.5 & 5.3 $\pm$ 0.5  & 11.7 $\pm$ 0.8 \\
HAIS \cite{chen2021hierarchical}      & 1.6 $\pm$ 0.6  & 3.5  $\pm$ 0.8 & 1.0 $\pm$ 0.2    & 2.3 $\pm$ 0.4  \\ 
\midrule
\Approach & \textbf{10.6 $\pm$ 0.7} & \textbf{19.8 $\pm$ 1.4} & \textbf{13.2 $\pm$ 0.9} & \textbf{24.8 $\pm$ 1.3 } \\ 
\bottomrule
\end{tabular}
\label{tab:compare_sota}
\end{table}

\begin{table*}[t]
\small
\setlength{\tabcolsep}{1pt}
\caption{Comparison of \Approach~and the strong baselines on S3DIS.}
\vspace{-10pt}
\centering
\begin{tabular}{lcccccc}
\toprule
    &   \multicolumn{3}{c}{$K=1$}    & \multicolumn{3}{c}{$K=5$}\\
    \cmidrule(lr){2-4}\cmidrule(lr){5-7}
            & mCov  & mPre     & mRec & mCov  & mPre     & mRec \\ 
\midrule
DyCo3D \cite{he2021dyco3d}         & {13.5 $\pm$ 2.1}& {2.9 $\pm$ 1.0}  & {4.1 $\pm$ 1.4}   & {14.5 $\pm$ 1.3}  & {3.1 $\pm$ 0.5}  & {4.1 $\pm$ 1.4} \\
PointGroup\cite{jiang2020pointgroup} & 12.9 $\pm$ 2.8 & 4.6 $\pm$ 1.4 & 3.8 $\pm$ 1.3 & 13.7 $\pm$ 0.8  & 4.6 $\pm$ 0.6 & 3.8 $\pm$ 0.8 \\
HAIS \cite{chen2021hierarchical}      & 4.6 $\pm$ 1.2  &  \textbf{8.1 $\pm$ 0.9}  & 3.9 $\pm$ 1.3  & 5.0 $\pm$ 1.9  &  \textbf{11.8 $\pm$ 2.0} & 4.1 $\pm$ 0.4 \\ 
\midrule
Ours & \textbf{17.8 $\pm$ 1.5} & 7.0 $\pm$ 0.4 & \textbf{8.5 $\pm$ 1.7} & \textbf{20.2 $\pm$ 2.1} & 10.8 $\pm$ 1.3 & \textbf{12.2 $\pm$ 1.8}     \\
\bottomrule
\end{tabular}
\label{tab:compare_sota_s3dis_1shot}
\vspace{-10pt}
\end{table*}

\begin{figure*}[t]
  \centering
  \includegraphics[width=1.0\linewidth]{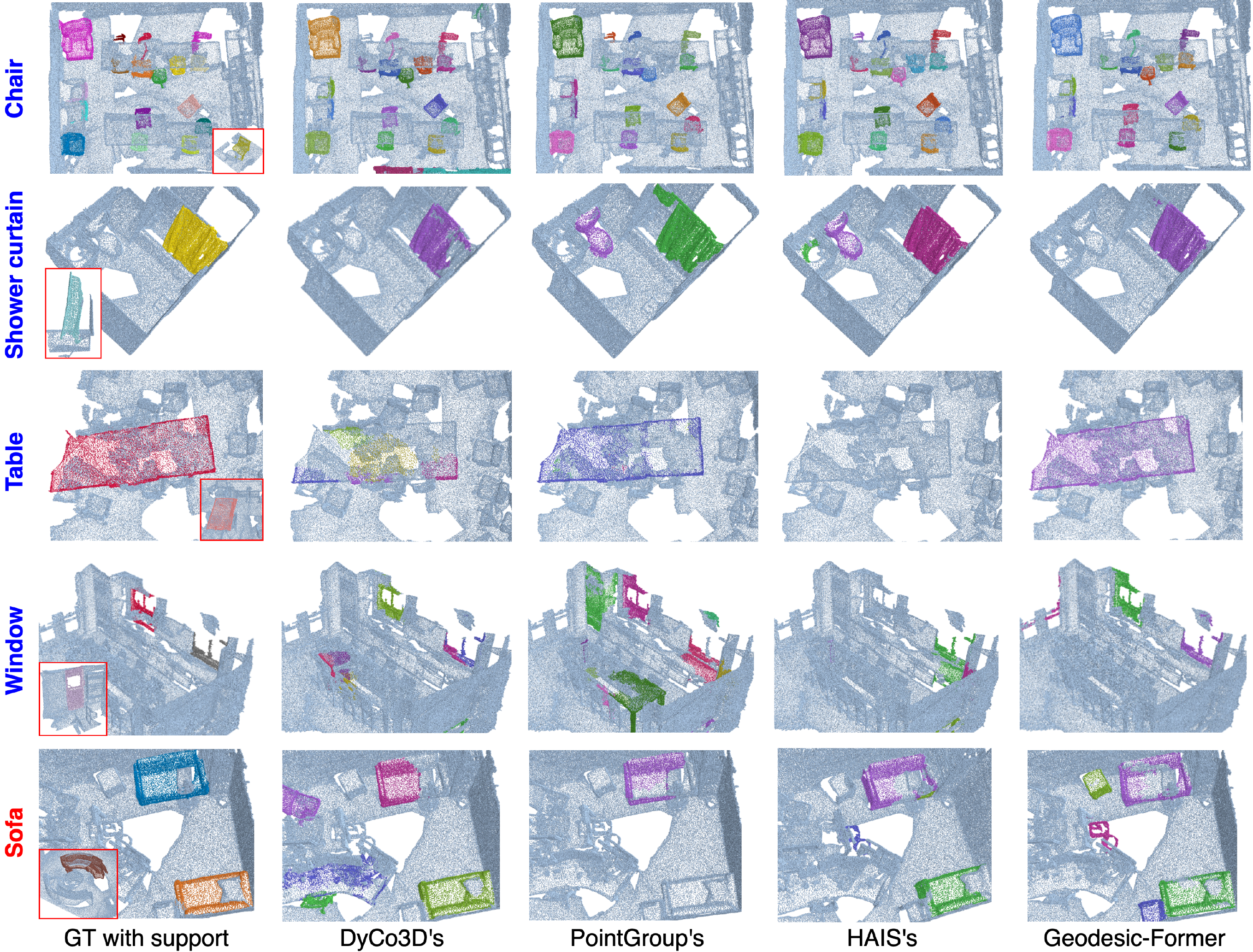}
  \vspace{-20pt}
   \caption{Qualitative results of \Approach~and the strong baselines on the ScannetV2 dataset. Each row shows an example of the query scene with its GT mask and the support scene with its GT mask (the smaller red-border box) on the first column. The name of the support class is on the left next to GT. 
   }
   \label{fig:quali}
\end{figure*}

\subsection{Comparison with Prior Work}

Since there is no prior work on 3DFSIS, we adapt three state-of-the-art (SOTA) approaches on 3DIS: DyCo3D \cite{he2021dyco3d}, PointGroup \cite{jiang2020pointgroup}, and HAIS \cite{chen2021hierarchical} to the few-shot setting for comparing with our approach. The adapted version of DyCo3D is exactly our baseline as depicted in Fig.~\ref{fig:baseline}. 
We apply the cosine similarity filter to all methods to remove irrelevant proposals after the clustering stage and the other modules are kept exactly the same as in their original papers. 
The similarity thresholds for these methods are carefully fine-tuned to achieve the best performance for a fair comparison, i.e. 0.95, 0.9, and 0.8 for DyCo3D, PointGroup, and HAIS, respectively.
Notably, the set aggregation module in HAIS requires another statistical class-specific radius to aggregate fragments into larger components. We calculate this radius based on the support scenes and then apply it to the query scene. 

Tab.~\ref{tab:compare_sota} and Tab.~\ref{tab:compare_sota_s3dis_1shot} show the comparison results on the S3DIS and ScannetV2 datasets, respectively. 
For ScannetV2, HAIS performs worst among the four, probably due to the sensitive class-specific radius in its set aggregation module. \Approach~consistently outperforms all of them by a large margin in all metrics, i.e., +4.4 for one shot and +6.8 for five shots in the mAP. Moreover, our method is more robust across different runs where the standard variations of mAP and AP50 are only 0.7 and 1.4, respectively, compared with 2.0 and 3.1 of the second-best DyCo3D's. 
For S3DIS, \Approach~outperforms others with a significant margin, i.e. in mCov and mRec, about +4 for one shot and +7 for five shots. HAIS's results are slightly better than ours in mPre due to its strict threshold to get high precision but low recall rate.

\subsection{Qualitative Results} 

\begin{figure*}[ht!]
  \centering
  \includegraphics[width=0.9\linewidth]{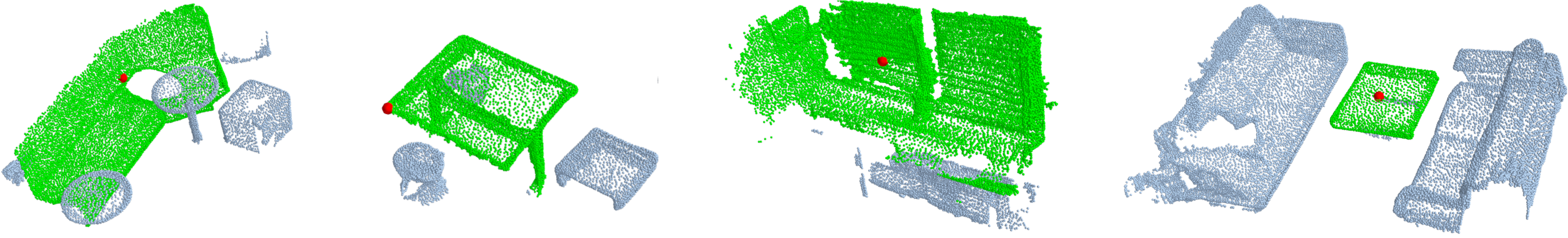}
   \caption{Representative examples of computed geodesic distance. For each image, the green points is the top reachable geodesic-distance nearest neighbors of the red point.}
   \label{fig:geodesic}
   \vspace{-10pt}
\end{figure*}


Fig.~\ref{fig:quali} shows the qualitative results of our approach and others on ScannetV2. For the training class ``chair'' shown on row 1, all approaches perform well. For the test classes (rows 2-5), there are differences in the segmentation results. \Approach~outperforms others in the hard cases such as in the thin object (``show curtain'' - row 2), in the big object (``table'' - row 3), and in the incomplete object (``window'' - row 4). These examples demonstrate the strong capability of our approach when handling objects to various extent thanks to the transformer decoder and the geodesic distance embedding. However, \Approach~mis-segments the sofa-stool as sofa due to their similar appearance (row 5).

Also, Fig.~\ref{fig:geodesic} illustrates the quality of the computed geodesic distance. For each red point, we visualize the top reachable geodesic-distance nearest neighbors (green points) and unreachable points (gray points) which have \textit{infinite} geodesic distance. It justifies that the geodesic distance helps distinguish objects much better than Euclidean distance.

\section{Discussion and Conclusion}
\label{sec:conclusion}

\textbf{Discussion.} We have succeeded in applying our approach on a lower number of shots only, i.e., 1 and 5 shots. For a higher number of shots ($K>5$), the improvement is insignificant due to the simple averaging operation. The study on how to aggregate features from multiple supports in a 3D point cloud to leverage their geometric structure would be an interesting research topic. 

\myheading{Conclusion.} In this work, we have introduced the new few-shot 3D point cloud instance segmentation task and have proposed the \Approach~-- a new geodesic-guided transformer with dynamic convolution to address it. Extensive experiments have been conducted on the newly introduced splits of ScannetV2 and S3DIS datasets showing that our approach achieves robust and significant performance gain on both datasets from the very strong baselines adapted from the state-of-the-art approaches in 3D instance segmentation, i.e., +4.4 for one shot and +6.8 for five shots in mAP on ScannetV2; +4.3 for one shot and +5.7 for five shots in mCov on S3DIS. We hope that our proposed problem, datasets, and approach could facilitate future work in this direction.



%
%
\bibliographystyle{splncs04}
\bibliography{egbib}
\end{document}